\documentclass[sigconf]{acmart}

\usepackage{amsmath,amssymb,amsfonts}
\usepackage{algorithmic}
\usepackage{graphicx}
\usepackage{multirow}
\usepackage{todonotes}
\AtBeginDocument{%
  \providecommand\BibTeX{{%
    \normalfont B\kern-0.5em{\scshape i\kern-0.25em b}\kern-0.8em\TeX}}}

\setcopyright{acmcopyright}
\copyrightyear{2018}
\acmYear{2018}
\acmDOI{10.1145/3535508.3545535}

\acmConference[Woodstock '18]{Woodstock '18: ACM Symposium on Neural
  Gaze Detection}{June 03--05, 2018}{Woodstock, NY}
\acmBooktitle{Woodstock '18: ACM Symposium on Neural Gaze Detection,
  June 03--05, 2018, Woodstock, NY}
\acmPrice{15.00}
\acmISBN{978-1-4503-XXXX-X/18/06}

\copyrightyear{2022}
\acmYear{2022}
\setcopyright{acmcopyright}\acmConference[BCB '22]{13th ACM International Conference on Bioinformatics, Computational Biology and Health Informatics}{August 7--10, 2022}{Northbrook, IL, USA}
\acmBooktitle{13th ACM International Conference on Bioinformatics, Computational Biology and Health Informatics (BCB '22), August 7--10, 2022, Northbrook, IL, USA}
\acmPrice{15.00}
\acmDOI{10.1145/3535508.3545535}
\acmISBN{978-1-4503-9386-7/22/08}

\begin{document}

\title[Modeling Long-term Dependencies and Short-term Correlations with Temporal Attention Networks for Health Prediction]{Modeling Long-term Dependencies and Short-term Correlations in Patient Journey Data with Temporal Attention Networks for Health Prediction}

\author{Yuxi Liu}
\authornote{Corresponding author}
\affiliation{%
  \institution{College of Science and Engineering, Flinders University}
  \city{Adelaide}
  \state{South Australia}
  \country{Australia}
}
\email{liu1356@flinders.edu.au}

\author{Zhenhao Zhang}
\affiliation{%
  \institution{College of Life Sciences, Northwest A\&F University}
  \city{Yangling}
  \state{Shaanxi}
  \country{China}
}
\email{zhangzhenhow@nwafu.edu.cn}

\author{Antonio Jimeno Yepes}
\affiliation{%
  \institution{School of Computing Technologies, RMIT University}
  \city{Melbourne}
  \state{Victoria}
  \country{Australia}
}
\email{antonio.jose.jimeno.yepes@rmit.edu.au}

\author{Flora D. Salim}
\affiliation{%
  \institution{School of Computer Science and Engineering, UNSW}
  \city{Sydney}
  \state{New South Wales}
  \country{Australia}
}
\email{flora.salim@unsw.edu.au}

\renewcommand{\shortauthors}{Y. Liu, et al.}

\begin{abstract}
Building models for health prediction based on Electronic Health Records (EHR) has become an active research area. EHR patient journey data consists of patient time-ordered clinical events/visits from patients. Most existing studies focus on modeling long-term dependencies between visits, without explicitly taking short-term correlations between consecutive visits into account, where irregular time intervals, incorporated as auxiliary information, are fed into health prediction models to capture latent progressive patterns of patient journeys. We present a novel deep neural network with four modules to take into account the contributions of various variables for health prediction: i) the \textit{Stacked Attention} module strengthens the deep semantics in clinical events within each patient journey and generates visit embeddings, ii) the \textit{Short-Term Temporal Attention} module models short-term correlations between consecutive visit embeddings while capturing the impact of time intervals within those visit embeddings, iii) the \textit{Long-Term Temporal Attention} module models long-term dependencies between visit embeddings while capturing the impact of time intervals within those visit embeddings, iv) and finally, the \textit{Coupled Attention} module adaptively aggregates the outputs of \textit{Short-Term Temporal Attention} and \textit{Long-Term Temporal Attention} modules to make health predictions. Experimental results on MIMIC-III demonstrate superior predictive accuracy of our model compared to existing state-of-the-art methods, as well as the interpretability and robustness of this approach. Furthermore, we found that modeling short-term correlations contributes to local priors generation, leading to improved predictive modeling of patient journeys.
\end{abstract}



\maketitle

\section{Introduction}
Developing deep learning (DL)-based models based on EHR data for predicting future health events or situations such as demands for health services has become an active research area, due to its potential to enable and facilitate care providers to take appropriate mitigating actions to minimize risks and manage demand. Examples of successful applications include early diagnosis prediction \cite{choi2016doctor}, disease risk prediction \cite{ma2018risk}, disease progression modeling and intervention recommendation \cite{pham2017predicting}.

Although the existing DL-based models have achieved promising results in many health prediction tasks, significant barriers remain when applying them for modeling EHR patient journey data. Raw EHR patient journey data has its own characteristics, such as temporal, multivariate, heterogeneous, irregular, and sparse nature, etc \cite{cheng2016risk}. This paper focuses mainly on the multivariate and temporal nature of EHR patient journey data, which contain inherent relationships at multiple levels and scales, notably, \textit{(i)} short-term correlations between consecutive clinical visits -- how every visit relates to each other in a short period and \textit{(ii)} long-term dependencies between clinical visits -- how each visit relates to the rest visits in the complete EHR patient journey. Integrated modeling of all these relationships is required to achieve more accurate predictions. However, to our knowledge, previous studies on EHR patient journey modeling have largely ignored the effect of \textit{(i)}. 

The following example is based on the publicly available MIMIC-III dataset \cite{johnson2016mimic}.

\begin{figure*}[!htb]
        \centering
        \includegraphics[width = 1.0\linewidth]{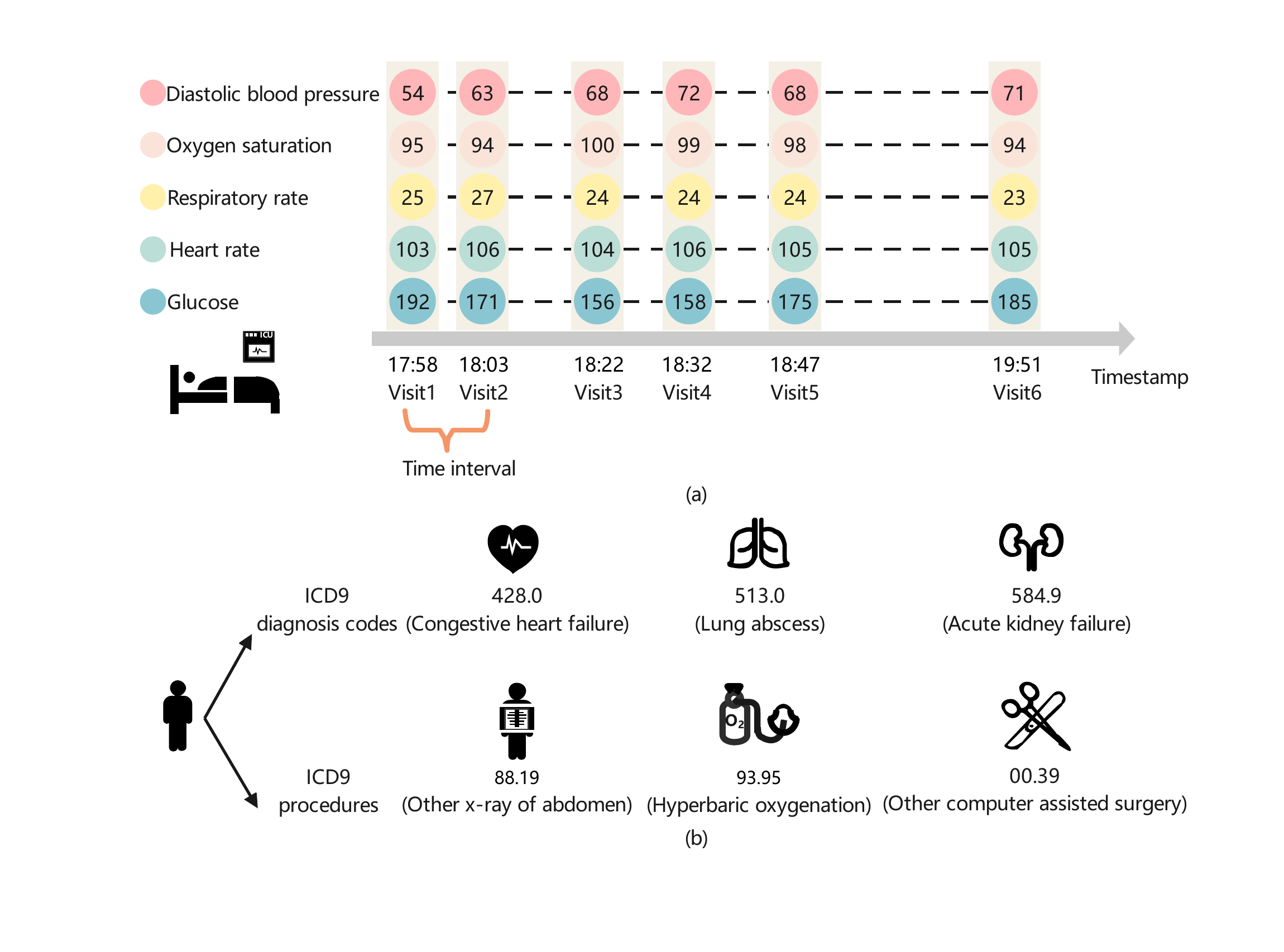}
        \caption{\textbf{Illustration of clinical time series. The time intervals between consecutive visits vary significantly.}}
        \label{fig:ATTANet_Overview}
\end{figure*}
The physician conducts the necessary lab tests for a patient at each visit during his/her intensive care unit (ICU) stays. This usually involves multiple clinical events for a patient that occur at the same point in time or within a short period. These clinical events are strongly related to a patient's health status, particularly in lab tests associated with a series of vital sign measurements (e.g., heart rate, blood glucose, respiratory rate. As shown in Figure 1a). The short-term correlations are how every visit relates to each other in a short period. The benefit of modeling these correlations is that such a consideration can contribute to local priors generation for improving predictive modeling of patient journeys. For example, when a patient is in a certain health status (i.e., 'severe' or 'healthier' status), certain values of a vital sign (e.g., blood glucose) within neighboring visits (e.g., consecutive visits) are likely to correlate to each other, and within those visits, there is a set of vital signs that are most relevant to that health status (i.e., local priors).

The modeling of accumulated patient journey data, i.e., a set of time-ordered clinical visits associated with a series of clinical events, forms a chain of data enabling researchers and health care providers to capture the long-term trends in patient health status. Besides, a patient's transition to a given health status also depends on his/her personal history of past clinical events, such as the previous disease diagnosis and procedures. Accordingly, both previous ICD9 diagnosis codes and procedures should be taken into consideration (As shown in Figure 1b). For example, when predicting the risk of mortality for patients, we should automatically include learning of the impact of a patient journey over the previous 48 hours on the prognosis. That is, using data from the first 48 hours of an ICU stay to predict in-hospital mortality for patients. In this sense, a predictive model should learn the context of a given patient journey through tracking and capturing the complex dynamic of clinical events and their interactions over time (i.e., long-term dependencies).

Clinical visit timestamps carry important information about the underlying patient journey dynamics, which allow us to talk about the chronologies (timing and order) of visits. In practice, it is important to consider the timing and order of visits to learn the context of a given patient's journey accurately. Usually, existing works construct sequential models by using recurrent neural networks (RNNs), and mainly model long-term dependencies between visits in patient journey data \cite{esteban2016predicting, jagannatha2016structured, choi2016doctor, pham2017predicting, suo2017multi}. 

In this paper, we propose a novel deep neural network with a modular structure, referred to as \textit{TAttNet} (Temporal Attention Networks) hereafter, to jointly tackle the above issues. The design of \textit{TAttNet} is inspired by ideas in processing sentences in documents from natural language processing (NLP). A patient journey is treated as a document and a clinical visit as a sentence. By learning the context of a given patient journey, \textit{TAttNet} is able to pay attention to both critical indicative previous clinical events, even though they happened a long time ago, and local clinical events between consecutive visits. Meanwhile, \textit{TAttNet} takes the time interval into consideration.
In addition, the architecture of \textit{TAttNet} provides interpretability of the model decisions, which is required to support its decisions, which is critical in machine learning EHR prediction systems.

We validate \textit{TAttNet} on the mortality prediction task from a publicly available EHR dataset, on which our method outperforms the baseline attention models by large margins. Further quantitative and qualitative analysis of the learned attentions shows that our method can also provide richer interpretations that align well with the views of relevant literature and medical experts.
\section{Related Work}
Recently, researchers have shown an increased interest in health prediction tasks. Recurrent neural networks such as LSTM \cite{hochreiter1997long}, GRU \cite{cho2014learning} and GRU-D \cite{che2018recurrent} have been used widely for EHR patient journey data modeling as representative deep learning models. Examples of successful applications include heart failure prediction \cite{maragatham2019lstm, choi2017using}, and comorbidity prediction and patient similarity analysis \cite{ruan2019representation}.
The performance of such sequential models is limited, especially for understanding the timing of visits, because the RNN architecture only contains recurrence (i.e., the order of sequences/visits). In reality, for example, measurements (vital signs) in clinical visits are commonly acquired with irregular time intervals (as shown in Figure 1a). To take the time interval into consideration, T-LSTM \cite{baytas2017patient} is proposed, which assumes that the clinical information may decay if there is a time interval between two consecutive visits. In other words, T-LSTM assumes that the more recent visits are more important than previous visits in general on health prediction tasks. T-LSTM transforms time intervals into weights and uses them to adjust the memory passed from previous moments in the Long Short Term Memory (LSTM). Based on this assumption, RetainEX \cite{kwon2018retainvis}, Timeline \cite{bai2018interpretable}, and ATTAIN \cite{zhang2019attain} are proposed, which focus on the provision of time-aware mechanisms for improving the predictive strength of RNNs. Their success makes combining RNNs with a time-aware mechanism module become the mainstream method in modeling EHR patient journey data, and they inspire the design principle of later works like ConCare \cite{ma2020concare} and HiTANet \cite{luo2020hitanet}.

It is worth noting that the work of HiTANet \cite{luo2020hitanet} was with a similar motivation to the above studies but with a different emphasis and method. HiTANet emphasizes that the previous clinical information should not be decayed because latent progressive patterns of patient journeys are non-stationary, where a patient's health condition can be better or worse at different timestamps. HiTANet designs a Time-aware Transformer to handle irregular time intervals of visits. It encodes time intervals into time vector representations and embeds them into visits first, and the Transformer model \cite{vaswani2017attention} then takes time vector representations as a part of the inputs and generates the hidden state representation of those visits. There is still much work to be done to achieve more adaptive time-aware mechanisms and desirable patient journey modeling simultaneously.

In addition, in real hospital scenarios, successful health prediction application requires not only the predictive strength of DL-based models but also model interpretability, which is essential to building trust in predictive models. To take the interpretability of predictive models into consideration, the innovative and seminal work of \cite{choi2016retain} developed a new DL-based model named RETAIN, which adapts two RNNs into an end-to-end interpretable network for health prediction tasks. The structure of RETAIN includes two RNNs, at visit-level and variable-level of patient contextual information respectively, to implement the attention mechanism, thereby capturing the influential past clinical visits and significant clinical variables within those clinical records. Results obtained from the RETAIN model often take an interpretable representation that is able to retain the semantics of each visit and variable while highlighting their influence on the target prediction by corresponding weights. Motivated by the successful application of RETAIN, attention-based DL-models are widely used in health prediction tasks. Examples of downstream applications include diagnosis prediction \cite{ma2017dipole}, mortality prediction \cite{sha2017interpretable}, and disease progression modeling \cite{bai2018interpretable}.

Compared to previous research, our work has a similar purpose but with a different emphasis and method. Our purpose is to construct an end-to-end interpretable network by combining deep neural networks and attention mechanisms together and mainly model EHR patient journey data. Our work emphasizes the use of \textit{Short-Term Temporal Attention} and \textit{Long-Term Temporal Attention} modules to separately model the short-term correlations and long-term dependencies in patient journey data. Compared to these prior works \cite{choi2016retain, ma2017dipole, lee2018diagnosis, zhang2020inprem, ma2020concare}, this provides richer patient journey representations, which are generated from long-term dependencies between visits and short-term correlations between consecutive visits, both promoting accurate modeling of patient journeys.

By rethinking our proposed method, the intuitions behind \textit{Short-Term Temporal Attention} and \textit{Long-Term Temporal Attention} modules can be seen as capitalizing on the main strength of a convolutional neural network (CNN) and a multilayer perceptron (MLP). On the one hand, \textit{Long-Term Temporal Attention} module builds upon an MLP to effectively model long-term dependencies between clinical visits. On the other hand, \textit{Short-Term Temporal Attention} module utilizes a CNN directly to fill in the gap of MLP in capturing short-term correlations between consecutive clinical visits. Another notable advantage of \textit{Short-Term Temporal Attention} module is that an interpretable attention map is given after the training, which shows the weights learned adaptively for variables that can describe how the importance of previous values of variables varies over time intervals and understand which variables are paid more attention to, during each visit for the target prediction. It is also worth noting that the generated two types of representations based on the above modules are simultaneously tuned together by designing a \textit{Coupled Attention} module to make health predictions.

Furthermore, it is worth bearing in mind that our \textit{TAttNet} starts from a \textit{Stacked Attention} module, and it strengthens the deep semantics in clinical events within each patient journey and generates visit embeddings. One notable advantage of this module is that an interpretable attention map is obtained after the training, which gives valuable information about the target variables on how much they are correlated to each other. Moreover, since it can learn asymmetric relationships, the attention map tells us which variables precede the others in terms of multivariate clinical time series forecasting.
\section{Methods}
\begin{figure*}[!htb]
        \centering
        \includegraphics[width = 1.0\linewidth]{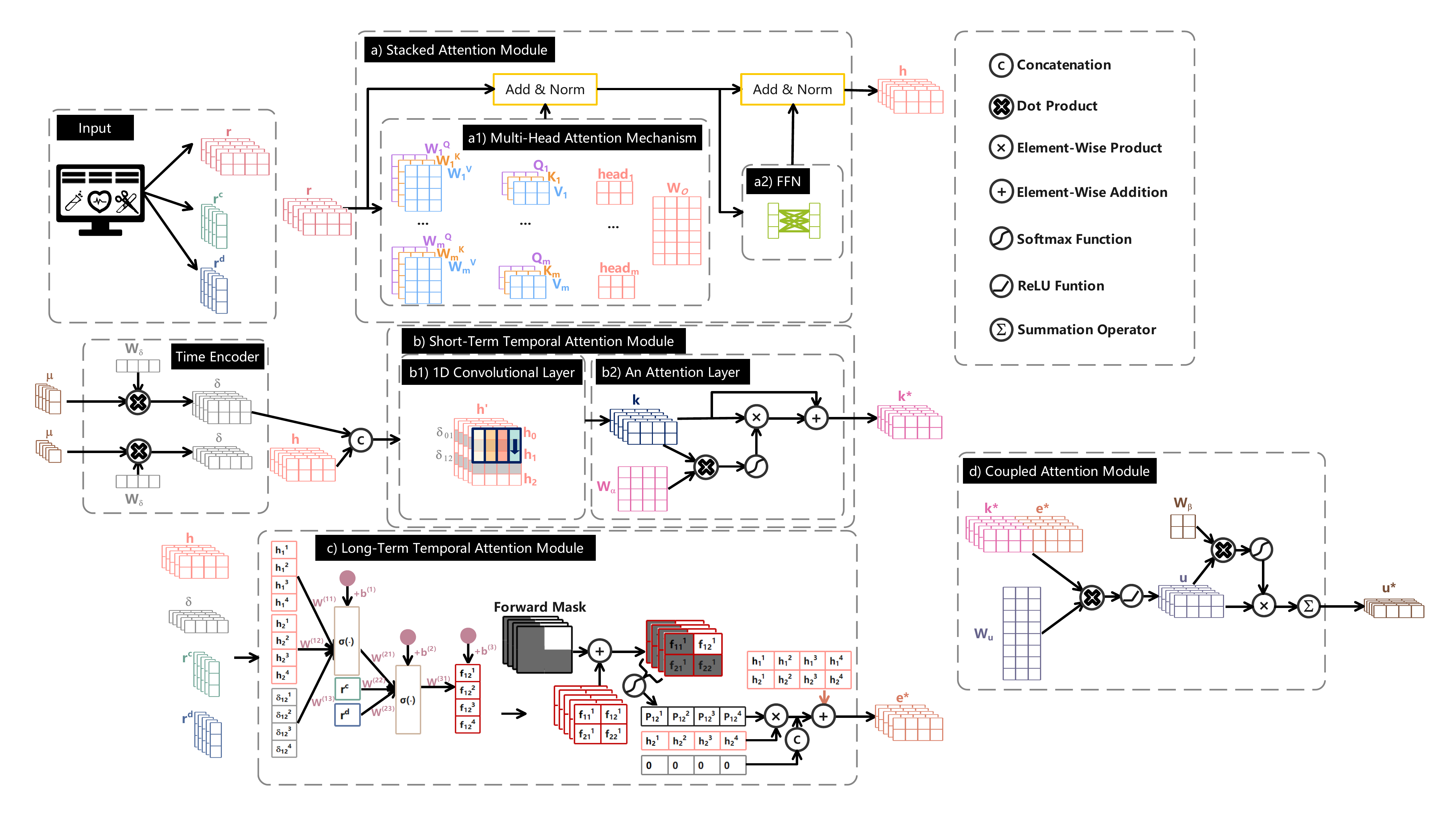}
        \caption{\textbf{Architecture of the proposed \textit{TAttNet}.}}
        \label{fig:TAttNet_Overview}
\end{figure*}
In this section, we introduce the proposed \textit{TAttNet} by discussing the basic notations of health prediction tasks first. Next, we detail four important modules of \textit{TAttNet}. Finally, we present how to use \textit{TAttNet} for health prediction tasks.
\subsection{Basic Notations}
\subsubsection{EHR Patient Journeys}
Let $P$ denote the set of all patient journeys, and $|P|$ is the number of patient journeys. Each patient journey $p \in P$ can be defined as follows:
\begin{equation}
\begin{split}
\label{eq:1}
r =
\left\{
\begin{matrix}
r_{11} & \cdots & r_{1T} \\
\vdots  & \ddots & \vdots \\
r_{N1} & \cdots & r_{NT}
\end{matrix}
\right\}, r^{c} = \{0, 1\}^{g_{c}}, r^{d} = \{0, 1\}^{g_{d}},
\end{split}
\end{equation}
where $r$ denotes clinical time series, which consists of the number of $T$ records and $N$ is the number of time-variant features in each record. Both $r^{c}$ (previous ICD9 diagnosis codes) and $r^{d}$ (ICD9 procedures) are time-invariant features. $g_{c}$ and $g_{d}$ are the number of unique ICD9 diagnosis codes and unique ICD9 procedures, respectively.
\subsubsection{The Temporal Nature of EHR Patient Journeys}
Let $\mu_{i}$ and $\mu_{j}$ denote the timestamp of records $r_{i}$ and $r_{j}$. $\delta_{ij}$ denotes the time interval between records $r_{i}$ and $r_{j}$, i.e., $\delta_{ij}$ = $\mu_{j}$ - $\mu_{i}$ ($i < j$).

\subsection{TAttNet architecture}
The architecture of the proposed \textit{TAttNet} is shown in Figure 2. In the following subsections, we provide the inner design of \textit{TAttNet} in greater detail.
\subsubsection{Stacked Attention Module}
In this subsection, we develop a \textit{Stacked Attention} module to strengthen the deep semantics in clinical events within each patient journey and generate visit embeddings (see Figure 2a). This module is mainly composed of the multi-head attention mechanism \cite{vaswani2017attention} whereby it contains multiple self-attention layers working in a parallel pattern. A self-attention layer includes three elements: a set of key-query pairs and values. The key-query pairs are used to compute the inner dependency weights, then used to update the values. Compared to a single self-attention layer, one advantage of the multi-head attention mechanism is that it enhances the attention layer with multiple representation subspaces. After obtaining the refined representation of each position by the multi-head attention mechanism, we add a feed-forward network (FFN) sub-layer.

Based on $r$, the generated representations $r^{MHA}$ can be defined as follows:
\begin{equation}
\begin{split}
\label{eq:3}
r^{MHA} = MultiHeadAttention(r) = \\ W_{o}[head_{1}(r) \oplus head_{2}(r) \oplus \cdots \oplus head_{m}(r)],
\end{split}
\end{equation}
where $head_{m}$ and $\oplus$ are the m-th attention head and concatenation operation, respectively. $W_{o} \in \mathbb{R}^{N \times mN}$ is a learnable parameter. For simplicity, we use $r^{MHA}$ to denote $MultiHeadAttention(r)$. In the following, we provide the details of all implementations.

First, we compute the attention weight ($\xi$) in Eq. (3) that determines how much each feature will be expressed at this certain feature.
\begin{equation}
\begin{split}
\label{eq:4}
\xi_{1}, \xi_{2},..., \xi_{N} = Softmax(\frac{Q^{\top} \cdot K_{1}}{\sqrt{d_{K}}}, \frac{Q^{\top} \cdot K_{2}}{\sqrt{d_{K}}},..., \frac{Q^{\top} \cdot K_{N}}{\sqrt{d_{K}}}),
\end{split}
\end{equation}
where $Q$ and $K_{i}$ denote query vector and key vector respectively, $d_{K}$ is the dimension of $K_{i}$, and the scaled dot product is used as the attention function \cite{vaswani2017attention}. $V_{i}$ denotes value vector. We use the way of \cite{wang2019r} to project input vectors into query, key and value spaces to obtain $Q$, $K_{i}$ and $V_{i}$ separately. The formula can be defined as follows:
\begin{equation}
\begin{split}
\label{eq:5}
Q, K_{i}, V_{i} = W_{Q} \cdot r^{\top}, W_{K} \cdot r^{\top}_{i}, W_{V} \cdot r^{\top}_{i},
\end{split}
\end{equation}
where $W_{Q}$, $W_{K}$ $\in$ $\mathbb{R}^{d_{K} \times T}$ and $W_{V} \in \mathbb{R}^{T \times T}$ are the projection matrices, which are learnable parameters. Each $head_{m}$ has its own projection matrix.

Second, we use $v_{i}$ and $\xi_{i}$ to compute $head_{m}$, which is obtained by using the weighted sum of $V_{i}$. The weights are calculated by applying the attention function to all key-query pairs.
\begin{equation}
\begin{split}
\label{eq:6}
head_{m}(r) = \sum^{N}_{i = 1} \xi_{i} \cdot V^{\top}_{i}.
\end{split}
\end{equation}

Based on Eq. (3-5), each $head_{m}$ is obtained by letting $r$ attend to all the feature positions so that any feature interdependencies between $r$ and $r_{i}$ can be captured.

After obtaining the refined new representation of each position by the multi-head attention mechanism, we use the way of \cite{vaswani2017attention} to add a feed-forward network sub-layer, which is used to transform the features non-linearly. The formula is defined as follows: $FFN(r^{MHA}) = W_{2} \cdot max(W_{1} \cdot r^{MHA} + b_{1}, 0) + b_{2}$. We also employ a residual connection \cite{he2016deep} around each of the two sub-layers, followed by layer normalization \cite{ba2016layer}. For simplicity, we use $h$ to denote the output of $FFN(r^{MHA})$. Last, the output of this module is $h \in \mathbb{R}^{N \times T}$.
\subsubsection{Learning Time Feature Embedding}
Before implementing \textit{Short-Term Temporal Attention} and \textit{Long-Term Temporal Attention} modules, we develop a Time-Encoder to embed $\delta$ and $h$ into the same feature space. The formula can be defined as below:
\begin{equation}
\begin{split}
\label{eq:2}
\delta_{enc} = W_{\delta} \cdot \delta + b_{\delta},
\end{split}
\end{equation}
where $W_{\delta} \in \mathbb{R}^{N \times 1}$ and $b_{\delta} \in \mathbb{R}^{N \times 1}$ are learnable parameters.
\subsubsection{Short-Term Temporal Attention Module}
In this subsection, we develop a \textit{Short-Term Temporal Attention} module to model short-term correlations between consecutive visit embeddings while capturing the impact of time intervals within those visit embeddings (see Figure 2b). The module consists of a customized 1D convolutional layer (i.e., kernel size=3, stride=2) and an attention layer. One notable advantage of the module is that an interpretable attention map is given after the training, which shows the weights learned adaptively for variables (clinical events) that can describe how the importance of previous values of variables varies over time intervals.

First, we embed time intervals into visit embeddings. Hence, $h = [h_{1}, h_{2}, \cdots, h_{T}]$ is expanded into $h^{\prime} = [h_{0}, \delta_{01}, h_{1}, \delta_{12}, h_{2}, \delta_{23}, h_{3}, \cdots\\, h_{T - 1}, \delta_{(T - 1)T}, h_{T}] \in \mathbb{R}^{N \times (2T + 1)}$. Note that $h_{0}$ is a zero vector. Second, we apply 1D convolution operation only over the horizontal dimension. To take time intervals into consideration, we specifically use a combination of $N$ kernels $\{W^{(j)}\}^{N}_{j = 1}$, and each kernel has kernel size 3 and stride 2. E.g., $h^{\prime}_{t: t + 2}$ denotes the concatenation of visit embeddings $h_{t-1}$, $h_{t}$ and time intervals $\delta_{(t-1)t}$ from $h_{t}^{\prime}$ to $h_{t + 2}^{\prime}$. A kernel $W^{(j)} \in \mathbb{R}^{3 \times 1}$ is naturally applied on the window of $h^{\prime (j)}_{t: t + 2}$ to produce a new feature $k_{(t+1)/2}^{(j)} \in \mathbb{R}$ with the ReLU activation function as follows:
\begin{equation}
\begin{split}
\label{eq:7}
k_{(t+1)/2}^{(j)} = ReLU(h_{t: t + 2}^{\prime (j)} \cdot W^{(j)} + b^{(j)}),
\end{split}
\end{equation}
where $b^{(j)} \in \mathbb{R}$ is a bias term and ReLU(k) = max(k, 0). This kernel is applied to each possible window of values in the whole description $\{h^{\prime (j)}_{1: 3}, h^{\prime (j)}_{3: 5}, \cdots, h^{\prime (j)}_{2T - 1: 2T + 1}\}$ to generate a feature map $k^{(j)} \in \mathbb{R}^{T}$ as follows: $k^{(j)} = [k^{(j)}_{1}, k^{(j)}_{2}, \cdots, k^{(j)}_{T}]$, $j = 1, 2, \cdots, N$. We can see that each kernel produces a feature. Since we have $N$ kernels, the final vector representation of a patient journey can be obtained by concatenating all the extracted features, i.e., $k \in \mathbb{R}^{N \times T}$. Last, an attention layer is applied to produce an attention weight $\alpha \in \mathbb{R}^{N \times T}$ and the final feature representation $k^{*}$ is obtained as follows:
\begin{equation}
\begin{split}
\label{eq:8}
\alpha = softmax(W_{\alpha} \cdot k + b_{\alpha}), \\
k^{*} = \alpha \odot k + k,
\end{split}
\end{equation}
where $W_{\alpha} \in \mathbb{R}^{N \times N}$ and $b_{\alpha} \in \mathbb{R}^{N \times 1}$ are learnable parameters. $\odot$ denotes element-wise multiplication. The generated attention weight $\alpha \in \mathbb{R}^{N \times T}$ encodes the causative and associative relationships of clinical events within each clinical visit embedding.
\subsubsection{Long-Term Temporal Attention Module}
In this subsection, we develop a \textit{Long-Term Temporal Attention} module to model long-term dependencies between visit embeddings while capturing the impact of time intervals within those visit embeddings (see Figure 2c). This module is specifically designed to learn the contextual information from each patient journey (i.e., the number of $T$ visit embeddings), whereby it generates an attention weight matrix $P^{\cdot}_{\cdot j} \in \mathbb{R}^{N \times T}$ ($j = 1, 2, \cdots, T$) that encodes the causative and associative relationships between the clinical events of j-th visit embedding and other visit embeddings. Mathematically, the module can be formalized as follow:
\begin{equation}
\begin{split}
\label{eq:10}
&f(h_{i}, h_{j}, \delta_{ij}) = \sigma (W^{(11)} h_{i} + W^{(12)} h_{j} + W^{(13)} \delta_{ij} + b^{(1)}), \\
&f(h_{i}, h_{j}, \delta_{ij}, r^{c}, r^{d}) = W^{(31)} \sigma (W^{(21)} f(h_{i}, h_{j}, \delta_{ij}) + W^{(22)} r^{c} \\ & + W^{(23)} r^{d} + b^{(2)}) + b^{(3)}, \\
&f^{\prime} (h_{i}, h_{j}, \delta_{ij}, r^{c}, r^{d}) = f(h_{i}, h_{j}, \delta_{ij}, r^{c}, r^{d}) + M^{fw}_{ij},
\end{split}
\end{equation}
where $h_{i}$ and $h_{j}$ denote the i-th and j-th visit embeddings of $h$. All $W$ and $b$ are learnable parameters. $\sigma(\cdot)$ is an activation function.

According to Eq. (9), we can see that the proposed module first computes $h_{i}$, $h_{j}$ and $\delta_{ij}$ to obtain the $f(h_{i}, h_{j}, \delta_{ij})$. It then builds the relationship between the obtained $f(h_{i}, h_{j}, \delta_{ij})$ and $r^{c}$, $r^{d}$ and generates the $f(h_{i}, h_{j}, \delta_{ij}, r^{c}, r^{d})$. Last, it incorporates a forward mask $M^{fw}_{ij}$ into the $f(h_{i}, h_{j}, \delta_{ij}, r^{c}, r^{d})$ and generates the $f^{\prime} (h_{i}, h_{j}, \delta_{ij}, r^{c}, r^{d})$.

The benefit of incorporating the mask operation \cite{peng2020self} (i.e., $M^{fw}_{ij}$) is that such a consideration can help the module learn the context of a patient journey according to the order of their clinical visits. That is, when computing each visit, the module only pays attention to all its previous visits. Specifically, a forward mask $M^{fw}$ is incorporated as follows:
\begin{equation}
\label{eq:11}
M^{fw}_{ij} =
  \begin{cases}
    0,       & i < j \\
    -\infty,  & otherwise.
  \end{cases}
\end{equation}

In order to generate an attention weight matrix (i.e., $P^{\cdot}_{\cdot j}$), a softmax function is applied to the $[f^{\prime}(h_{i}, h_{j}, \delta_{ij}, r^{c}, r^{d})]^{T}_{i = 1}$ and thus obtain the probability distribution $p(g|h, h_{j})$. $g$ is an indicator of which feature is important to $h_{j}$. A large $P^{l}_{ij}$, $P^{l}_{ij} \triangleq p(g_{l} = i|h, h_{j})$ means that $h_{i}$ contributes important information to $h_{j}$ on the $l$-th feature dimension. For simplicity, we ignore the subscript $l$ if it does not cause any confusion.

The $e_{j}$ is obtained by traversing all $h_{i}$ as follows:
\begin{equation}
\label{eq:14}
e_{j} = \sum^{T}_{i = 1} P^{\cdot}_{ij} \odot h_{i}, j = 1, 2, \cdots, T,
\end{equation}
where $e_{j}$ is the weighted average of sampling a visit embedding according to its importance. We obtain $e = [e_{1}, e_{2}, \cdots, e_{T}] \in \mathbb{R}^{N \times T}$ from Eq. (11). We add $e$ to $h$ to obtain the final feature representation $e^{*}$, i.e., $e^{*}$ = $e$ + $h$.
\subsubsection{Coupled Attention Module}
Through the aforementioned design, \textit{TAttNet} has been able to model long-term dependencies and short-term correlations in patient journey data. Meanwhile, it takes the time interval into consideration. The two types of representations ($k^{*}$ and $e^{*}$), generated based on \textit{Short-Term Temporal Attention} and \textit{Long-Term Temporal Attention} modules, should be simultaneously tuned together. In this subsection, we develop a \textit{Coupled Attention} module to adaptively couple both representations (see Figure 2d). Given the obtained representations $k^{*}$ and $e^{*}$, an overall representation $u$ is generated as follows:
\begin{equation}
\label{eq:15}
u = ReLU(W_{u} \cdot [k^{*}_{t}, e_{t}^{*}] + b_{u}),
\end{equation}
where $W_{u} \in \mathbb{R}^{d_{u} \times 2N}$ and $b_{u} \in \mathbb{R}^{d_{u}}$ are learnable parameters. $d_{u}$ is the dimension of the overall representation $u$. The rectified linear unit is defined as ReLU(x) = max(x, 0). Note that max() applies element-wise to vectors.

Next, an attention layer is applied to $u$ to obtain a final representation $u^{*}$ as follows:
\begin{equation}
\label{eq:16}
\beta = softmax(u \cdot W_{\beta} + b_{\beta}), \\
u^{*} = \sum^{T}_{i = 1} \beta_{i} \odot u_{i},
\end{equation}
where $W_{\beta} \in \mathbb{R}^{T \times T}$ and $b_{\beta} \in \mathbb{R}^{T}$ are learnable parameters. $\odot$ denotes element-wise multiplication.
\subsection{Health Prediction}
The output of subsection 4.5, a final representation $\{u^{*}_{p}\}^{|P|}_{p = 1}$, is fed into a Softmax output layer to obtain the prediction probability $\hat{y}_{p}$:
\begin{equation}
\begin{split}
\label{eq:16}
\hat{y}_{p} = softmax(W_{y} \cdot u^{*}_{p} + b_{y}),
\end{split}
\end{equation}
where $W_{y}$ and $b_{y}$ are learnable parameters. The cross-entropy between the ground truth $y_{p}$ and the prediction probability $\hat{y}_{p}$ is used to calculate the loss. Thus, the objective function of health prediction is the average of cross entropy:
\begin{equation}
\begin{split}
\label{eq:17}
\mathcal{L(\theta)} = - \frac{1}{|P|} \sum^{|P|}_{p = 1} (y^{\top}_{p} \cdot log(\hat{y}_{p}) + (1 - y_{p}) ^{\top} \cdot log(1- \hat{y}_{p})),
\end{split}
\end{equation}
where $\theta$ is the parameter of \textit{TAttNet} and $|P|$ is the total number of patient journeys.
\section{Experiments}
In this section, we report the experimental results on a publicly available EHR dataset to demonstrate the effectiveness of the proposed method. After the discussion of the experimental setting, we first compare our approach with attention-based DL methods. Moreover, we analyze the effectiveness and interpretability of our approach by an ablation study and case studies.
\subsection{Experimental Setup}
\subsubsection{Datasets and Tasks}
We validate the performance of our method on health risk prediction tasks from the publicly available Medical Information Mart for Intensive Care (MIMIC-III) dataset \cite{johnson2016mimic}, for both the prediction accuracy and prediction robustness. The data were normalized on the basis of the literature \cite{harutyunyan2019multitask}.

MIMIC-III is one of the largest publicly available ICU datasets, comprising 38,597 distinct patients and a total of 53,423 ICU stays. We utilize clinical times series data (e.g., heart rate, respiration rate) and previous ICD9 diagnosis codes as well as ICD9 procedures as inputs. The prediction tasks here are two binary classification tasks, 1) in-hospital mortality (48 hours after ICU admission): to evaluate ICU mortality based on the data from the first 48 hours after ICU admission, 2) in-hospital mortality (complete hospital stay after ICU admission): to evaluate ICU mortality based on the data during the complete hospital stay.
\subsubsection{Baselines}
To fairly evaluate the effectiveness of the proposed \textit{TAttNet}, we implement the following attention-based DL methods:\\
\begin{enumerate}
    \item GRU$_{\alpha}$ is the basic GRU \cite{cho2014learning} with a location-based attention mechanism \cite{ma2017dipole}.
    \item RETAIN \cite{choi2016retain} (NeurIPS 2016) is an interpretable deep learning model for health prediction. It purposes the use of a two-level attention mechanism (based on two RNNs), which could enhance both the performance and interpretability of the model.
    \item Dipole \cite{ma2017dipole} (SIGKDD 2017) has a similar purpose with RETAIN, but with a different method. It uses a bidirectional RNN and incorporates a three-level attention mechanism.
    \item Transformer$_{e}$ is the encoder of the Transformer \cite{vaswani2017attention} (NeurIPS 2017); in the final step, we use to flatten and feed-forward networks to make health predictions.
    \item INPREM \cite{zhang2020inprem} (SIGKDD 2020) can be seen as a combination of a linear part and a non-linear part. It mainly uses three attention mechanisms to fill in the gap of a linear model in learning the non-linear relationship of features from EHR patient journey data.
    \item T-LSTM \cite{baytas2017patient} (SIGKDD 2017) is an improved LSTM based approach by modifying gate information to model the time decay.
    \item RetainEX \cite{kwon2018retainvis} is built upon RETAIN to learn weights for clinical visits and events while capturing the impact of time intervals within those visits. It mainly handles previous clinical visits by enabling time decay.
\end{enumerate}
\subsubsection{Implementation Details \& Evaluation Strategies}
We perform all the baselines and \textit{TAttNet} with Python v3.7.0. For each task, we randomly split the datasets into training, validation, and testing sets in a 75:10:15 ratio. The validation set is used to select the best values of parameters. Binary outcomes were evaluated with the area under the receiver operating characteristic curve (AUROC) and the area under the precision-recall curve (AUPRC). We repeat all the approaches ten times and report the average performance.

We use class weight in CrossEntropyLoss for a highly imbalanced dataset. This is achieved by placing an argument called 'weight' on the CrossEntropyLoss function. This approach offers an effective way of alleviating the problem of being highly imbalanced.

In order to evaluate the interpretability of our \textit{TAttNet}, the following steps were taken: \textit{(i)} we randomly select a case (i.e., one patient journey) from the MIMIC-III dataset \cite{johnson2016mimic}. \textit{(ii)} we input imputed values for the missing items of this case on the basis of literature \cite{harutyunyan2019multitask}. \textit{(iii)} we use the case as the input of the trained \textit{TAttNet}, and the mask operation is applied to both \textit{Stacked Attention} and \textit{Short-Term Temporal Attention} modules. Altogether, this makes the dimensionality of the data input to \textit{TAttNet} appropriate and avoids taking imputed items into consideration in the target prediction.
\subsection{Performance Evaluation}
Table 1 shows the performance of all methods on the MIMIC-III dataset. The results indicate that \textit{TAttNet} significantly and consistently outperforms other baseline methods. E.g., for the mortality prediction of MIMIC-III (48 hours after ICU admission), \textit{TAttNet} achieves the highest AUROC with 0.9076, one standard deviation of 0.002. Similarly, \textit{TAttNet} achieved the highest AUPRC with 0.6326, one standard deviation of 0.016.

We find that all methods demonstrated good prediction robustness for lengthy EHR patient journeys. E.g., for the mortality prediction of MIMIC-III (complete hospital stay after ICU admission), RETAIN and Dipole achieved AUROCs of 0.8663 and 0.8884, respectively, which is an improvement over the prediction performance of their on 48 hours by roughly 2\%. In contrast, \textit{TAttNet} achieves the best AUROC/AUPRC scores amounting to 0.9538 and 0.7999.
\begin{table*}[htbp]
  \centering
  \caption{The AUROC/AUPRC scores of in-hospital mortality prediction with different observation windows for MIMIC-III dataset.}
    \begin{tabular}{cccccc}
    \toprule
    \multicolumn{2}{c}{MIMIC-III/Mortality Prediction} & \multicolumn{2}{c}{48 hours after ICU admission} & \multicolumn{2}{c}{complete hospital stay after ICU admission} \\
    \midrule
    \multicolumn{2}{c}{Metrics} & AUROC & AUPRC & AUROC & AUPRC \\
    \midrule
    \multirow{8}[0]{*}{Methods} & GRU$_{\alpha}$ & 0.8339(0.005) & 0.6248(0.006) & 0.8406(0.008) & 0.6213(0.010) \\
          & RETAIN & 0.8429(0.003) & 0.4848(0.008) & 0.8663(0.010) & 0.6014(0.013) \\
          & Dipole & 0.8685(0.002) & 0.5357(0.004) & 0.8884(0.005) & 0.5180(0.007) \\
          & Transformer$_{e}$ & 0.8116(0.008) & 0.6046(0.010) & 0.8260(0.014) & 0.5923(0.017) \\
          & INPREM & 0.8699(0.008) & 0.5207(0.012) & 0.8807(0.008) & 0.5399(0.018) \\
          & T-LSTM & 0.8743(0.002) & 0.4354(0.024) & 0.9153(0.010) & 0.7367(0.031) \\
          & RetainEX & 0.8737(0.005) & 0.4668(0.016) & 0.9240(0.005) & 0.7496(0.017) \\
          & \textit{TAttNet} & \textbf{0.9076(0.002)} & \textbf{0.6326(0.016)} & \textbf{0.9538(0.003)} & \textbf{0.7999(0.015)} \\
    \bottomrule
    \end{tabular}%
  \label{tab:addlabel}%
\end{table*}%
\subsection{Ablation Study}
We now need to examine the effectiveness of different modules of our method. To this end, we conduct an ablation study on the datasets. To determine whether the developed modules improve the prediction performance, we present four variants of \textit{TAttNet} as follows:

\textit{TAttNet}$_{\alpha}$: \textit{TAttNet} without \textit{Stacked Attention}.

\textit{TAttNet}$_{\beta}$: \textit{TAttNet} without \textit{Short-Term Temporal Attention}.

\textit{TAttNet}$_{\gamma}$: \textit{TAttNet} without \textit{Long-Term Temporal Attention}.

\textit{TAttNet}$_{\delta}$: \textit{TAttNet} without \textit{Coupled Attention}.

We present the ablation study results in Table 2. We find that \textit{TAttNet} outperforms \textit{TAttNet}$_{\beta}$ (i.e., without the \textit{Short-Term Temporal Attention} module). It indicates that modeling of short-term correlations between consecutive visits contributes to local priors generation, for improving predictive modeling of patient journeys. \textit{TAttNet} also outperforms \textit{TAttNet}$_{\delta}$ (i.e., without the \textit{Coupled Attention} module), which demonstrates that adaptively aggregates two types of representations are generated by \textit{Short-Term Temporal Attention} and \textit{Long-Term Temporal Attention} modules which can produce enhanced prediction performance. The superior performance of \textit{TAttNet} than the \textit{TAttNet}$_{\alpha}$ verifies the efficacy of the \textit{Stacked Attention} module which can generate richer patient journey representations and improve the performance.
\begin{table*}[htbp]
  \centering
  \caption{Ablation performance comparison.}
    \begin{tabular}{cccccc}
    \toprule
    \multicolumn{2}{c}{MIMIC-III/Mortality Prediction} & \multicolumn{2}{c}{48 hours after ICU admission} & \multicolumn{2}{c}{complete hospital stay after ICU admission} \\
    \midrule
    \multicolumn{2}{c}{Metrics} & AUROC & AUPRC & AUROC & AUPRC \\
    \midrule
    \multirow{5}[0]{*}{Methods} & \textit{TAttNet}$_{\alpha}$ & 0.9005(0.002) & 0.6173(0.018) & 0.9109(0.002) & 0.6406(0.018) \\
          & \textit{TAttNet}$_{\beta}$ & 0.9036(0.002) & 0.6002(0.017) & 0.9360(0.006) & 0.7389(0.027) \\
          & \textit{TAttNet}$_{\gamma}$ & 0.8886(0.008) & 0.6228(0.010) & 0.9376(0.003) & 0.7511(0.011) \\
          & \textit{TAttNet}$_{\delta}$ & 0.9064(0.002) & 0.6243(0.010) & 0.9491(0.003) & 0.7930(0.011) \\
          & \textit{TAttNet} & 0.9076(0.002) & 0.6326(0.016) & 0.9538(0.003) & 0.7999(0.015) \\
    \bottomrule
    \end{tabular}%
  \label{tab:addlabel}%
\end{table*}%
\subsection{Case Study: Method Interpretability}
We validate the interpretability of our \textit{TAttNet} with random examples selected from the datasets, which are presented in Figures 3 and 4. The proposed \textit{TAttNet} enjoys good interpretability owing to both \textit{Stacked Attention} and \textit{Short-Term Temporal Attention} modules. Specifically, the \textit{Stacked Attention} module is able to provide an interpretable attention map after the training, which gives valuable information about the target variables on how much they are correlated to each other. The \textit{Short-Term Temporal Attention} module is able to provide an interpretable attention map after the training, which shows the weights learned adaptively for variables that can describe how the importance of previous values of variables varies over time intervals and understand which variables are paid more attention to during each visit for the target prediction.

\begin{figure}[!htb]
        \centering
        \includegraphics[width = 1.0\linewidth]{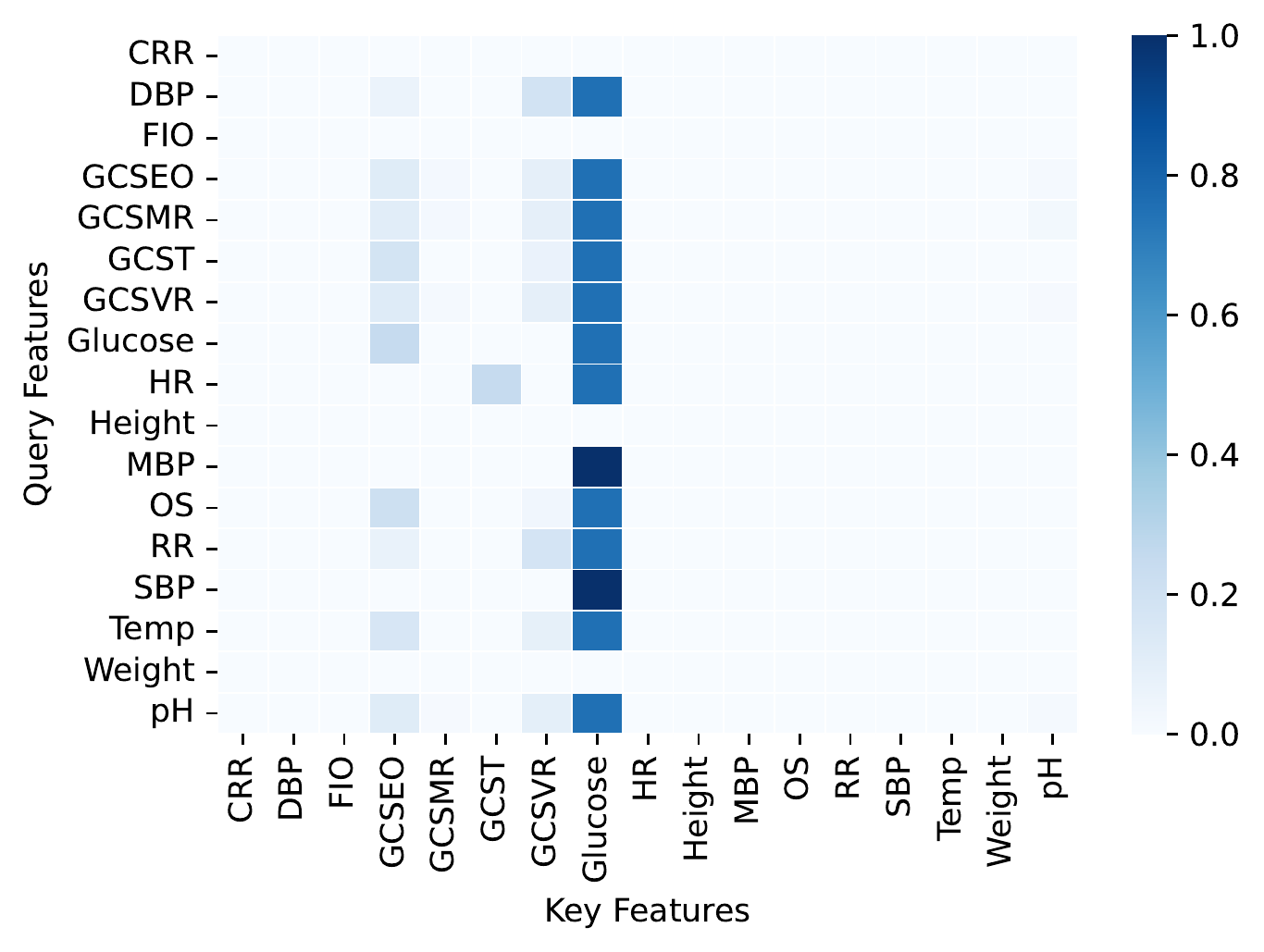}
        \caption{\textbf{The variable/feature importance of patient A with respect to the other variables/features. In patient A, diabetes mellitus is the primary disease, accompanied by a series of diabetes-induced diseases and other diseases.}}
        \label{fig:Stacked_Attention}
\end{figure}

Figure 3 shows the variable/feature importance of patient A who died after 48 hours of an ICU stay. The feature weights ranging from $0.0 \sim 1.0$ were calculated by the proposed \textit{Stacked Attention} module. The ordinates y-axis of the figure shows the Query features, and the abscissas x-axis shows the Key features. The boxes intensity of color in the shaded areas in Figure 3 shows how much each Key feature responds to the Query when a Query feature makes a query. Looking at Figure 3, it is apparent that most of the features related to the primary disease are more likely to respond to each other, represented by the feature weights of two matrices (See Figure 3 for details of the primary disease of patient A).

It is common medical knowledge that the glucose of a patient is strongly related to diabetes. As shown in Figure 3, in the box of Glucose-Glucose position, \textit{TAttNet} pays much more attention to the glucose in patient A. \textit{TAttNet} determines that there are relatively high associations between pH and glucose. This is highly consistent with the medical literature \cite{yoshida2018relationship, tang2000effects}. Moreover,~\textit{TAttNet} determines that there are relatively high associations between diastolic blood pressure (DBP), systolic blood pressure (SBP), and glucose. This result may be explained by the fact that patient A has both essential hypertension and diabetes. This finding broadly supports the work of other studies in this area linking essential hypertension with diabetes \cite{mancia2005association, zhao2017prevalence, lv2018association}. This finding also supports evidence from clinical observations \cite{henry2002impaired} that there are relatively high associations between abnormal glucose, blood pressure, and mortality. Furthermore, \textit{TAttNet} determines that there are relatively high associations between four Glasgow Coma Scale scores (i.e., GCSEO, GCSMR, GCST, GCSVR) and glucose. These are highly consistent with the medical literature \cite{kotera2014analysis}.

\begin{figure}[!htb]
        \centering
        \includegraphics[width = 1.0\linewidth]{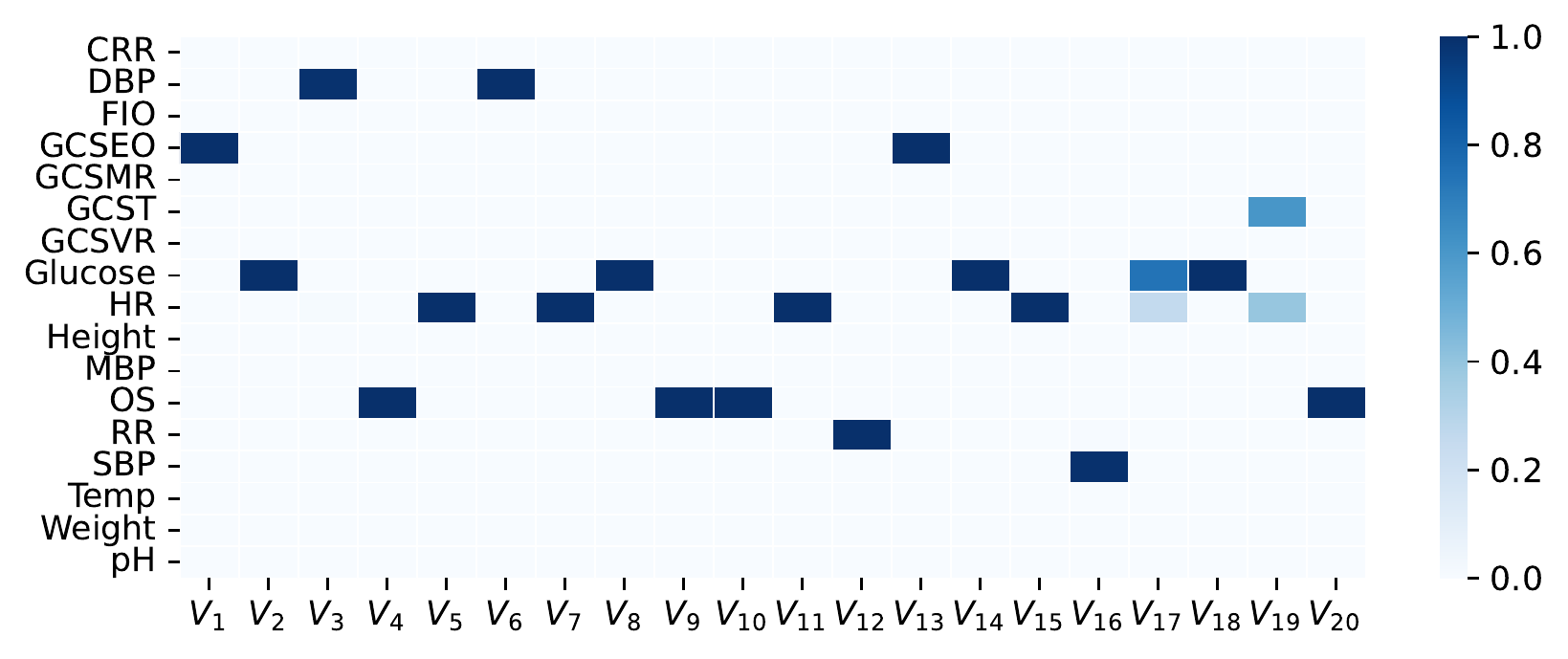}
        \caption{\textbf{The variable/feature importance of patient A with respect to each one of the visits (V1 to V20).}}
        \label{fig:Short-Term_Temporal_Attention}
\end{figure}
Figure 4 shows the variable/feature importance of patient A who died after 48 hours of an ICU stay (Note that both Figures 3 and 4 have used the same patient A). The feature weights ranging from $0.0 \sim 1.0$ were calculated by the proposed \textit{Short-Term Temporal Attention} module. The boxes intensity of color in the shaded areas in Figure 4 shows how much each feature responds to the target prediction when the module calculates a visit. Note that the module considers all visits of patient A (i.e., an entire journey of patient A), but the images are understandably truncated for visibility. We take the first 20 visits of patient A as an example for detailed discussion.

In Figure 4, there are observed changes in the importance of previous values of variables vary over time intervals (i.e., at different visits). E.g., in the box of Glucose-$V_{2}$ and Glucose-$V_{8}$ positions, \textit{TAttNet} pays much more attention to the glucose in patient A. After zooming on Visits 2 and 8 ($V_{2}$, $V_{8}$), we find that glucose had greater weight, contributing most to the corresponding visit. When querying the data of patient A from the MIMIC-III dataset, we found that glucose had a larger value of 608 mg/dL within Visit 2. Similarly, glucose had a larger value of 440 mg/dL within Visit 8. In reviewing the literature \cite{gunst2019glucose}, we found that the commonly-used standard of glucose in critically ill patients is below 180 mg/dL. Overall, these result provides reasonably consistent evidence of an association between abnormal glucose and mortality \cite{barr2007risk, itzhaki2017admission}.
\section{LIMITATIONS AND FUTURE WORKS}
Several limitations to this study need to be acknowledged. \textit{(i)} MIMIC-III database is a large medical dataset comprising all information relating to patients admitted to intensive care units. This work uses clinical times series data and previous ICD9 diagnosis codes as well as ICD9 procedures as inputs. Further investigation and experimentation into more informative details such as admission information and free text diagnosis are strongly recommended. \textit{(ii)} In our case studies, the scope of the experimental data was chosen based on the literature \cite{ozyurt2021attdmm}, i.e., the first 48 hours of an ICU stay and complete hospital stay after ICU admission. In terms of directions for future research, further work could explore patient-specific contextual information at various time points, such as 24 or 36 hours. It would be worthwhile to compare patient-specific contextual information at various time points, and the experimental results may offer different prediction probabilities and medical findings. \textit{(iii)} The lack of model uncertainty analysis in the results adds further caution regarding the generalisability of these findings. This paper argues that successful health prediction requires high predictive accuracy and higher-quality model interpretability. However, DL-based models are usually highly parameterized and thus may increase the uncertainty of their prediction results. In order to achieve high predictive accuracy, researchers often make many efforts to find the best parameters. Hence, the model results often depend on the customization of parameters, but the parameters are uncertain, achieved by analyzing different perspectives and using different optimal algorithms, even human experiences. Future studies should address the questions raised by the model parameter uncertainty (also known as model uncertainty). Altogether, this will build more trust in predictive models and thus enable the transition from academic research to clinical applications.
\section{CONCLUSIONS}
In this paper, we introduce a novel deep neural network named \textit{TAttNet} for health prediction tasks in EHR patient journey data mining. This prediction will enable and facilitate care providers to take appropriate mitigating actions to minimize risks and manage demand. We conduct experiments on a publicly available EHR dataset. \textit{TAttNet} yielded performance improvements in both AUROC and AUPRC over the state-of-the-art prediction methods. Furthermore, our \textit{TAttNet} also provides an interpretation as to the cause of individual patient predictions, allowing the user to take more important features into consideration in future investigations. The authors believe this may be useful for providing personalized estimates of outcome probabilities. A number of possible future studies using the same experimental setup are apparent such as hospital stays, re-admissions, and future diagnoses.


\end{document}